\documentclass[3p,times,authoryear]{elsarticle}
\usepackage{ecrc}
\volume{00}
\firstpage{1}
\journalname{Journal name}
\runauth{J. Chen}
\jid{ANM}
\makeatletter
\let\@dochead\@empty
\def\ps@pprintTitle{%
  \let\@oddhead\@empty
  \let\@evenhead\@empty
  \let\@oddfoot\@empty
  \let\@evenfoot\@empty}
\def\ps@headings{%
  \def\@oddhead{\parbox{\textwidth}{\itshape\footnotesize%
      \hfill\@runauth\hfill{\rm \thepage}}}%
  \def\@evenhead{\parbox{\textwidth}{\itshape\footnotesize%
      {\rm \thepage}\hfil\@runauth\hfil}}%
  \let\@oddfoot\@empty
  \let\@evenfoot\@empty}
\long\def\MaketitleBox{%
  \resetTitleCounters
  \def\baselinestretch{1}%
  \begin{center}%
  \ifx\@dochead\@empty\relax%
     \vspace*{2pc}%
  \else%
     \vspace*{3.5pc}%
   \@dochead%
     \par%
     \vspace*{1.75pc}%
   \fi%
   \def\baselinestretch{1}%
    {\strut\elsarticle@titlefont\@title\strut}\par\vskip18pt
    {\elsarticle@authorfont\elsauthors}\par\vskip10pt
    \footnotesize\itshape\elsaddress\par\vskip36pt
    \hrule\vskip12pt
    \ifvoid\absbox\else\unvbox\absbox\par\vskip10pt\fi
    \ifvoid\keybox\else\unvbox\keybox\par\vskip10pt\fi
    \hrule\vskip12pt
    \end{center}%
   \ifcase\jtype\or
    \vspace*{-20pt}%
   \or
   \or
    \vspace*{-20pt}%
   \fi}
\makeatother
\usepackage{amsfonts}
\usepackage{amssymb}
\usepackage{amsmath}
\usepackage{amsthm}
\usepackage[mathlines]{lineno}
\usepackage{graphicx}
\usepackage{hyperref}
\usepackage{mathtools}
\usepackage[figuresright]{rotating}
\usepackage{booktabs}
\usepackage{subcaption}
\usepackage{tabularx}
\usepackage{array}
\usepackage{cleveref}
\usepackage{color}
\newcommand{\R}{\mathbb{R}}
\newcommand{\E}{\mathbb{E}}
\newcommand{\Var}{\operatorname{Var}}

\newcommand{\KL}{D_{\mathrm{KL}}}
\newcommand{\X}{\mathcal{X}}
\newcommand{\Y}{\mathcal{Y}}

\newcommand{\Dn}{\mathcal{D}_n}

\newcounter{procedure}
\newenvironment{procedure}[1]
{\refstepcounter{procedure}\par\medskip
 \noindent\textbf{Procedure~\theprocedure\ (#1).}
 \begin{enumerate}}
{\end{enumerate}\medskip}
\theoremstyle{plain}

\newtheorem{lemma}{Lemma}[section]
\newtheorem{definition}{Definition}[section]

\theoremstyle{remark}
\newtheorem{remark}{Remark}[section]
\newtheorem{example}{Example}

\begin{document}
\begin{frontmatter}

\title{Diagnosing the Conditional-Mean Barrier in Scientific Machine-Learning Surrogates}

\author{Junfeng Chen}
\ead{majfchen@ust.hk}
\address{Department of Mathematics, The Hong Kong University of Science and Technology, Hong Kong, China}

\begin{abstract}
Many prediction tasks in computational science and engineering become one-to-many after coarse graining and partial observation. In such settings, deterministic surrogates trained by squared loss may learn a well-defined mathematical object—the conditional mean—while still missing the task-relevant variability in the underlying conditional law. In this work, we formulate this limitation as the \emph{conditional-mean barrier} and develop a diagnostic framework for identifying it in fitted scientific machine-learning surrogates. The framework combines residual-feature orthogonality and effect-size diagnostics to distinguish deterministic underfitting from irreducible conditional variability. We also make explicit a simple consequence of paired squared loss: stochastic outputs do not by themselves overcome the barrier, because the objective penalizes model variance and drives the predictor back to the conditional mean. The diagnosis therefore yields a modeling prescription: when residual variability matters, the loss must score richer features of the conditional law rather than a point prediction. Reproducible numerical studies on a controlled two-branch law and a two-scale Lorenz--96 closure problem show how the diagnostic identifies the barrier, how deterministic closures can suppress collective fluctuation statistics in rollout, and how a minimal likelihood-based stochastic-scale model can recover substantially more variability.
\end{abstract}


\end{frontmatter}

\section{Introduction}\label{sec:intro}

A common use of scientific machine learning is to discover, accelerate, or augment an expensive, unresolved, or partially observed component of a computation by a data-driven surrogate. In a typical supervised formulation,
one observes pairs \((x_i,y_i)\), chooses a model \(f_\theta\), and trains it
so that \(y_i\approx f_\theta(x_i)\). This template appears in learned
closures and subgrid models for resolved states
\citep{brunton2016discovering,rudy2017data,duraisamy2019turbulence}, in
surrogates for expensive solution operators
\citep{lu2021learning,kovachki2023neural}, and in inverse reconstruction from
limited measurements \citep{hansen2010discrete,benning2018modern}. Because it
is stable, interpretable, and easy to optimize, squared-loss training of a
deterministic surrogate often serves as the default starting point.

The complication is that the data pairs used to train such a surrogate need
not come from a single-valued map. After coarse graining or partial
observation, the same resolved descriptor can correspond to many possible
responses. A coarse state in a multiscale dynamical system does not uniquely determine the
unresolved forcing, because the hidden variables may carry memory and induce
stochastic effects \citep{chorin2000optimal,lu2017data}. A low-dimensional
descriptor of a heterogeneous medium can be compatible with many three-dimensional
microstructures and therefore with a range of effective properties
\citep{gommes2012microstructural,bostanabad2018computational}. A low-resolution image can have many plausible high-resolution completions
\citep{ledig2017photo,saharia2023image}. In probabilistic terms, the
conditional law \(P(Y\mid X=x)\) is then not a point mass.

Squared loss handles this ambiguity in a precise but limited way. Its population target is the conditional mean \(\E[Y\mid X]\) \citep[Sec.~2.4]{hastie2009elements}\citep[Sec.~1.5.5]{bishop2006pattern}. \textcolor{black}{This characterization is classical, and it is the starting point of our analysis:} a sufficiently expressive deterministic surrogate can be entirely successful as a squared-loss predictor while still discarding the spread, tails, or multimodality present in \(P(Y\mid X)\). In scientific computing, the discarded
variability may be exactly what a downstream calculation needs, for example
when a closure is rolled out autonomously and the quantity of interest is an
energy fluctuation, a flux statistic, a risk bound, or an ensemble spread.

The two-scale Lorenz--96 closure study in this paper provides a concrete instance. A local deterministic closure trained by least squares can reach close to the conditional mean. Yet, when inserted into a closed autonomous system, it strongly suppresses the variance of energy fluctuations, even though the system remains dynamically active and simple one-point statistics are nearly preserved. The issue is therefore not that the closure is poorly fitted to its one-step squared-loss objective. Rather, the rollout statistic is sensitive to conditional variability around that mean.

This creates a practical fork in model development. If the residual still
contains structure that can be explained by the chosen resolved variables,
then the deterministic surrogate has not finished its job: one should enlarge
the model class, improve the optimization, or collect better
data. If, however, the surrogate has reached the conditional mean and the
remaining residual variability is still too large for the scientific task,
then more deterministic capacity is the wrong remedy. The target has changed
from a point predictor to some feature of the conditional law \(P(Y\mid X)\).
The central question is how to distinguish these two cases in finite data.

\begin{figure}[t]
  \centering
  \includegraphics[width=0.8\linewidth]{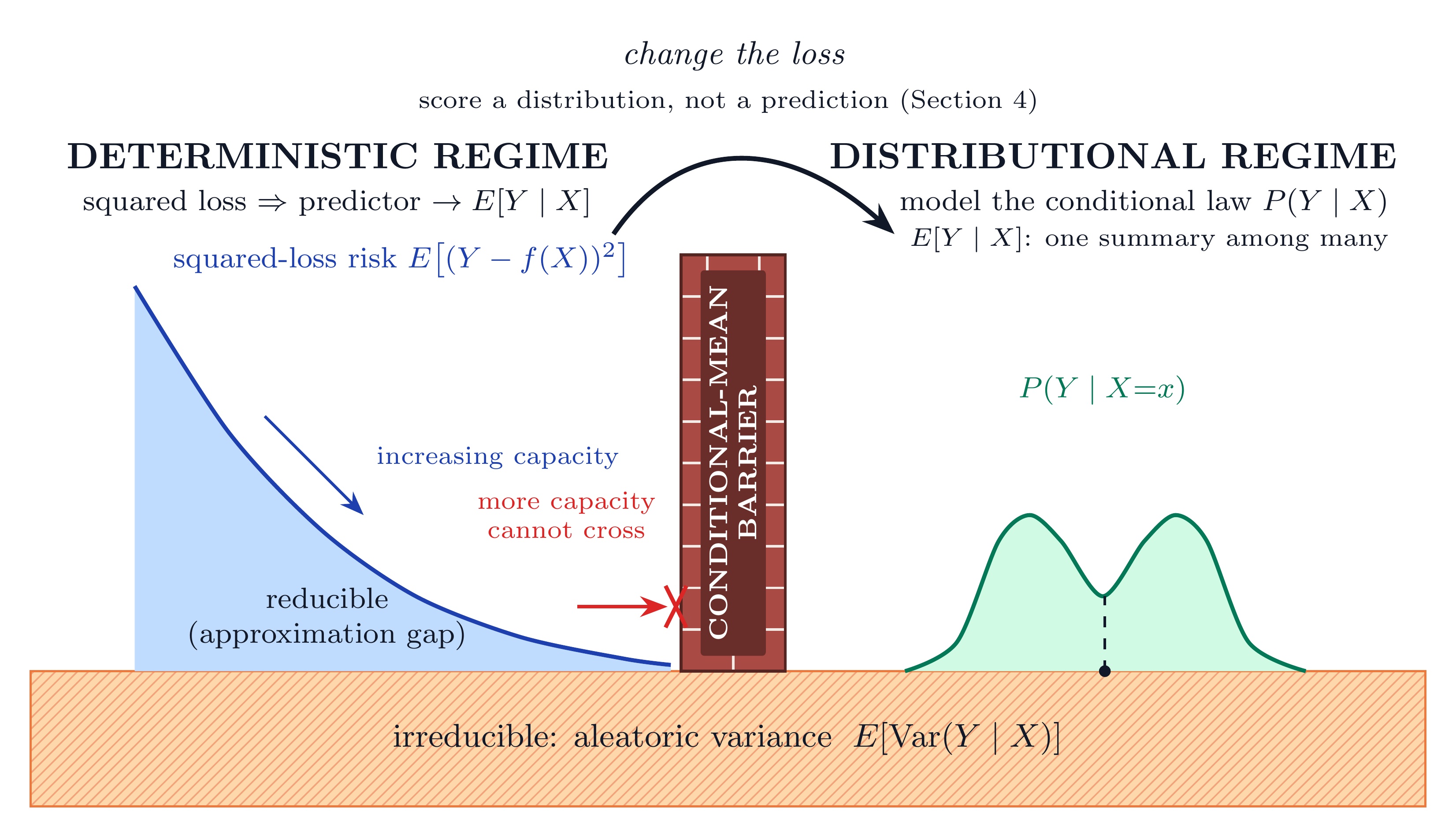}
  \caption{The conditional-mean barrier. In the deterministic regime
    (left), increasing model capacity drives the squared-loss risk
    $E[(Y-f(X))^2]$ down toward the irreducible aleatoric floor
    $E[\mathrm{Var}(Y\mid X)]$ as $f \to E[Y\mid X]$; the reducible
    approximation gap closes, but no squared-loss predictor can pass below
    the floor, and adding capacity---or latent randomness
    (Section~\ref{sec:distributional})---does not help. Crossing into the
    distributional regime (right) requires changing the loss to one that
    scores a distribution rather than a prediction, after which the object
    of inference is the conditional law $P(Y\mid X)$, of which the
    conditional mean is a single summary.}
  \label{fig:barrier}
\end{figure}

We call the second operating point the \emph{conditional-mean barrier}
(Figure~\ref{fig:barrier}): the fitted squared-loss surrogate has extracted the
deterministic signal available from the chosen resolved variables, but the
irreducible conditional variability remains scientifically relevant. This
paper develops the barrier as an operational diagnostic framework for
machine-learning surrogates in scientific computing. The framework first
locates the barrier by combining residual-feature orthogonality, effect
sizes, and the explained-variance ceiling. It then makes explicit why a
tempting repair remains misaligned, namely adding latent randomness while
continuing to train with squared loss, and finally points to distributional
objectives that score the feature of \(P(Y\mid X)\) required by the task.

The paper makes four contributions. 
\begin{itemize}
\item First, it formulates the conditional-mean barrier as \textcolor{black}{an operational} modeling fork in scientific machine learning: residual error may reflect either deterministic underfitting or conditional
variability irreducible relative to the chosen input \(X\).

\item Second, it develops a finite-data diagnostic rule that combines residual-feature orthogonality, statistical significance, effect sizes, and the residual mean-square to decide whether to refine a deterministic surrogate, stop, or change the loss; once deterministic structure has been exhausted, the stabilized residual mean-square estimates the aleatoric floor.

\item Third, it shows why stochastic outputs trained by paired squared loss do not overcome the barrier: the loss penalizes model variance and drives the predictor back to the conditional mean.

\item Fourth, it demonstrates the pipeline on a controlled two-branch law and a two-scale Lorenz--96 closure problem, where a deterministic mean closure attains reasonable one-step accuracy yet suppresses slow-energy fluctuations in autonomous rollout, and a minimal likelihood-based stochastic-scale model substantially reduces this suppression.
\end{itemize}

This diagnosis also fixes the role of the distributional objectives discussed later: they are not default replacements for deterministic regression, but post-diagnostic choices used when the scientific task requires more than the conditional mean.

The remainder of this paper is organized as follows. Section~\ref{sec:setup}
fixes notation and formulates the surrogate-learning dataset.
Section~\ref{sec:deterministic} develops the squared-loss limit, the
residual-feature diagnostic, and the explained-variance ceiling.
Section~\ref{sec:distributional} first explains why stochastic outputs trained
by paired squared loss still collapse to the conditional mean, then uses
negative log-likelihood (NLL) as a minimal distributional objective, and finally maps other post-diagnostic objectives to
scientific needs. Section~\ref{sec:experiments} reports the controlled
two-branch and two-scale Lorenz--96 studies, including a minimal
stochastic-scale intervention for fluctuation suppression.

\section{Setup and notation}\label{sec:setup}

We formulate the supervised learning problem through a pair of
random variables \((X,Y)\) with joint law \(P_{X,Y}\). The input \(X\)
represents the information made available to the model, such as a coarse-grained state, a local stencil, a partial observation, or another incomplete description of the underlying system. The output \(Y\) is the response to be inferred from that
information, such as an unresolved closure forcing, a global property, or a
reconstructed high-resolution field. In most
examples one may take \(\X\subseteq\R^d\) and \(\Y\subseteq\R^k\), but the same
notation also covers image-valued or function-valued outputs.
We write \(P_X\) and \(P_Y\) for the marginals and \(P_{Y\mid X}\) for the
conditional law of \(Y\) given \(X\). Such a law exists, for example, on
standard Borel spaces, which includes the Euclidean settings used below and
many separable function-space settings encountered in applications
\citep[Ch.~5]{kallenberg2002foundations}. Throughout, \(\E\) and \(\Var\)
denote expectation and variance under the joint law unless a subscript
indicates otherwise.

The joint law \(P_{X,Y}\) is not observed directly. Instead, one works with a
finite collection of paired samples
\begin{equation}\label{eq:dataset}
\Dn = \{(x_i, y_i)\}_{i=1}^{n},
\qquad
(x_i, y_i) \stackrel{\mathrm{iid}}{\sim} P_{X,Y}.
\end{equation}

We distinguish three data roles. The \emph{training set} is used to fit the
surrogate. An independent \emph{diagnostic set} is used to examine residual
structure and decide whether further deterministic refinement is warranted or
whether the fit has reached the conditional-mean barrier. A separate
\emph{test set}, when used, is reserved for final held-out evaluation and is
not used to choose the model or interpret the residual diagnostics. This
separation is essential for the residual-orthogonality tests developed in
Section~\ref{sec:deterministic}. Training residuals may be orthogonal, by
construction, to the features used during fitting; diagnostic residuals test
whether structure remains in directions the fitting procedure was not forced
to match. Unless otherwise stated, \(\Dn\) denotes the training set.

A learning procedure is specified by a hypothesis class and a loss function.
For a loss \(\ell\), the population risk is
\[
    \mathcal R(f) = \E[\ell(Y,f(X))],
\]
and the empirical risk replaces the expectation by an average over the
training data. Practical algorithms minimize empirical risk over a restricted
class, so approximation error, estimation error, optimization error, and
finite-sample rates remain important in applications
\citep[Ch.~2]{mohri2018foundations}. Here, we focus on a complementary issue: the loss function determines which population object the
model is trying to learn. Squared loss selects the conditional mean,
whereas likelihood-based training selects a distributional approximation to
the conditional law within the chosen model family. The diagnostic question is
therefore: when has a squared-loss model learned all detectable
deterministic signal in the chosen input \(X\), and when does the scientific
task require a different feature of \(P(Y\mid X)\)?

\section{Squared-loss surrogates and the conditional-mean barrier}
\label{sec:deterministic}

We now specialize to deterministic surrogates trained by squared loss. A fitted
map \(\hat f:\X\to\Y\) returns one output for each input, and therefore cannot
by itself represent a non-degenerate conditional law \(P_{Y\mid X=x}\). The
loss determines which point summary of that law is learned. Under squared
loss, the summary is the conditional mean. The aim of this section is to turn this classical population fact into an operational diagnostic: when has a deterministic scientific surrogate exhausted the signal available in the chosen input \(X\), and when does the residual error reflect irreducible conditional variability that cannot be represented by any point predictor?

The diagnostic rests on two projection facts. First, the squared-loss
population minimizer over all measurable functions of \(X\) is the conditional
mean \(m(X)=\E[Y\mid X]\). Second, at this minimizer the residual
\(Y-m(X)\) is orthogonal to every square-integrable function of \(X\). The
first fact identifies the target of deterministic squared-loss learning; the
second becomes the residual-feature diagnostic used to decide whether the
target has been reached in finite data.

\subsection{The conditional mean as an $L^{2}$ projection}
\label{subsec:cmean}

For notational simplicity we state the result for scalar \(Y\). The
vector-valued case follows componentwise under squared Euclidean loss. We recall the standard squared-loss characterization of conditional
expectation and its \(L^2\)-projection interpretation.

\begin{lemma}[Optimality of the conditional mean]\label{thm:cmean} Let \(Y\) be square-integrable. Among all measurable \(f:\X\to\R\) such that \(f(X)\) is square-integrable, the conditional mean \(m(X)=\E[Y\mid X]\) is a minimizer of
\begin{equation}\label{eq:cm-opt}
    \inf_{f \text{ measurable}}
    \E\bigl[(Y-f(X))^2\bigr],
\end{equation}
and is unique up to \(P_X\)-almost-sure equality. Equivalently, the map \(Y\mapsto \E[Y\mid X]\) is the orthogonal projection of \(Y\) onto the closed subspace \(L^2(\sigma(X))\subseteq L^2(P)\) of square-integrable, \(\sigma(X)\)-measurable random variables.
\end{lemma}

{\color{black}This is the standard \(L^2\)-projection characterization of conditional expectation; we omit the proof and refer to \citep[Sec.~2.4]{hastie2009elements}\citep[Sec.~1.5.5]{bishop2006pattern}. Writing \(W=Y-m(X)\) and using \(\E[W\mid X]=0\), the same argument gives the deterministic squared-loss decomposition}
\begin{equation}\label{eq:det-decomp}
\E\bigl[(Y-f(X))^2\bigr]
=
\E\bigl[(m(X)-f(X))^2\bigr]
+
\E\bigl[\Var(Y\mid X)\bigr],
\qquad
m(X)=\E[Y\mid X].
\end{equation}
The first term is the reducible approximation gap to the conditional mean.
The second is the irreducible conditional variability left after the best
deterministic squared-loss prediction has been made.

This decomposition is the starting point of the diagnostic framework. In the
population limit, a sufficiently expressive deterministic surrogate trained by
squared loss is searching for \(m(x)=\E[Y\mid X=x]\). Linear models in rich
feature maps, kernel methods \citep{steinwart2001influence,schaback2006kernel},
and neural networks \citep{hornik1991approximation,higham2019deep} differ in
how they parametrize this search and how they trade approximation against
estimation error; they do not differ in the population object selected by
squared loss. Once the approximation gap in \eqref{eq:det-decomp} has been
removed, the remaining floor \(\E[\Var(Y\mid X)]\) is not a failure of
deterministic optimization or capacity. It is the conditional spread left by
the information contained in \(X\).

Lemma~\ref{thm:cmean} says that the conditional mean is the optimal
deterministic summary under squared loss. It does not say that the conditional
mean is a typical or even plausible realization of \(Y\mid X=x\). This
distinction is sharpest in the following example.

\begin{example}[Two-branch conditional law]\label{ex:two-branch}
Let \(X\) have distribution \(P_X\) on \(\X\), and let
\(a,b:\X\to\R\) be measurable square-integrable functions with
\(a(x)\neq b(x)\) on a set of positive \(P_X\)-measure. Suppose that, for
each \(x\in\X\),
\[
Y\mid X=x
\sim
\tfrac12\delta_{a(x)}+\tfrac12\delta_{b(x)}.
\]
Then
\[
m(x)=\E[Y\mid X=x]=\tfrac12\bigl(a(x)+b(x)\bigr),
\]
which lies between the two branches and differs from both whenever the
branches differ. The squared-loss-optimal deterministic predictor therefore
returns a value that the conditional law \(P_{Y\mid X=x}\) never realizes at
those inputs. The conditional variance is
\[
\Var(Y\mid X=x)=\tfrac14\bigl(a(x)-b(x)\bigr)^2,
\]
strictly positive wherever the two branches separate.
\end{example}

This is the simplest form of the conditional-mean barrier. A deterministic
surrogate may be wrong because it has not yet learned the conditional mean, or
it may be right as a mean predictor while still missing the variability that
the application needs. The diagnostic below is designed to separate these two
situations.

\subsection{Approximating the conditional mean}
\label{subsec:basis}

In applications, \(m\) is approximated within a chosen hypothesis class. For a
linear surrogate in features \(\{\varphi_j\}_{j=1}^p\), this means
approximating \(m\) by \(\sum_j\beta_j\varphi_j\). Since
Lemma~\ref{thm:cmean} gives \(m\in L^2(P_X)\), the population approximation
question is whether the chosen features are rich enough in \(L^2(P_X)\).

\begin{definition}[Complete basis]\label{def:complete}
A countable family \(\{\varphi_j\}_{j=1}^{\infty}\subset L^2(P_X)\) is
\emph{complete} in \(L^2(P_X)\) if its linear span is dense in \(L^2(P_X)\);
equivalently, every \(g\in L^2(P_X)\) admits a sequence of linear combinations
of \(\{\varphi_j\}\) converging to \(g\) in \(L^2(P_X)\).
\end{definition}

Polynomial, Fourier, and wavelet bases are standard examples, and kernel or
neural-network classes provide other routes to rich approximation under
appropriate hypotheses \citep{steinwart2001influence,cybenko1989approximation,
hornik1991approximation,leshno1993multilayer}. The particular parametrization
is not the main issue here. The diagnostic question begins when richer
deterministic models no longer remove residual structure: is the remaining
error a missed component of the conditional mean, or conditional variability
around that mean?

\subsection{Diagnosing the conditional-mean barrier}
\label{subsec:diagnostics}

Lemma~\ref{thm:cmean} gives a population certificate for having reached the
conditional mean:
\[
    f(X)=m(X) \quad P_X\text{-a.s.}, \qquad m(X)=\E[Y\mid X],
\]
if and only if
\begin{equation}\label{eq:orth-char}
\E\bigl[(Y-f(X))\,g(X)\bigr]=0
\qquad
\text{for every } g\in L^2(\sigma(X)).
\end{equation}
The forward direction is the orthogonality part of the projection proof. The
reverse direction follows by taking \(g=m-f\). If
\(\{\psi_k\}_{k=1}^{\infty}\subset L^2(P_X)\) is complete, then it is enough,
at the population level, to test \eqref{eq:orth-char} against this family, by
density.

Our use of this classical certificate is operational. We combine it with
finite-data residual probes, effect sizes, and the residual mean-square to
answer a modeling question: has the fitted surrogate merely underfit the
conditional mean, or has it reached the conditional-mean barrier relative to
the chosen input \(X\)? These ingredients play different roles. Residual
moments and significance detect whether structure remains. Effect sizes
measure whether that structure is large enough to matter. The residual
mean-square then decides whether the remaining conditional variability is
acceptable for the scientific task.

\subsubsection{Residual-feature moments and statistical significance}

Let \(\{(x_i,y_i)\}_{i=1}^n\) be an independent diagnostic set, and write
\[
    \hat r_i = y_i-\hat f_n(x_i).
\]
For a diagnostic probe \(\psi\), define the population residual-feature
moment and its empirical counterpart
\[
    \mu_\psi(f)
    \coloneqq
    \E\left[\bigl(Y-f(X)\bigr)\psi(X)\right], \qquad 
    \hat\mu_\psi
    \coloneqq
    \frac1n\sum_{i=1}^n \hat r_i\,\psi(x_i).
\]
At the conditional mean, \(\mu_\psi(m)=0\) for every admissible probe. For testing the null moment condition \(\mu_\psi=0\), we use the
central-limit scaling
\begin{equation}\label{eq:T-stat}
T_n(\psi)
\coloneqq
\sqrt n\,\hat\mu_\psi
=
\frac1{\sqrt n}\sum_{i=1}^n \hat r_i\,\psi(x_i).
\end{equation}
Under the ideal null \(\hat f_n=m\) and standard moment
conditions, the central limit theorem gives
\[
T_n(\psi)
\quad\Longrightarrow\quad
N\left(0,\,
\E\bigl[\Var(Y\mid X)\psi(X)^2\bigr]\right).
\]
With
\[
    \hat s_\psi^2
    =
    \frac1n\sum_{i=1}^n
    \bigl(\hat r_i\psi(x_i)-\hat\mu_\psi\bigr)^2,
\]
we form the standardized statistic
\begin{equation}\label{eq:significance}
    t_n(\psi)
    =
    \frac{\sqrt n\,\hat\mu_\psi}{\hat s_\psi}.
\end{equation}

In practice the diagnostic is applied to a finite probe family
\[
    \Psi=\{\psi_1,\ldots,\psi_{q}\},
\]
rather than to a single probe in isolation. For each \(\psi\in\Psi\), we call it statistically \emph{significant} at family level \(\alpha\) if
\[
    |t_n(\psi)|>c_{\alpha,q}.
\]
A simple Bonferroni family-wise threshold is
\[
    c_{\alpha,q}
    =
    t_{1-\alpha/(2q),n-1}.
\]

Significance detects whether a residual trend is reliably different from zero.
It should not, however, be used as the modeling stopping rule. If the true
residual-feature moment is \(\mu_\psi(f)\neq 0\), then
\[
    t_n(\psi)
    =
    \sqrt n\,
    \frac{\mu_\psi(f)}
         {\sqrt{\Var((Y-f(X))\psi(X))}}
    + O_p(1).
\]
Thus, for any fixed nonzero moment, \(|t_n(\psi)|\) grows like \(\sqrt n\).
With enough diagnostic data, even a very small residual-feature moment can
become statistically significant. Significance therefore answers whether a
moment is reliably different from zero; by itself, it does not measure the
impact of correcting that moment on the squared-loss risk. The latter
quantity is the effect size introduced below.

\subsubsection{Effect size as removable residual structure}

The effect size measures the reduction in residual mean-square that would be
obtained by adding one diagnostic correction direction. Write
\[
    r_f \coloneqq Y-f(X), \qquad
    \mu_\psi(f) \coloneqq \E[r_f\psi(X)], \qquad
    \sigma_r^2(f) \coloneqq \E[r_f^2], \qquad
    v_\psi \coloneqq \E[\psi(X)^2].
\]
If the current predictor is corrected to \(f+a\psi\), its population
squared-loss risk becomes
\[
\E\bigl[(Y-f(X)-a\psi(X))^2\bigr]
=
\sigma_r^2(f)-2a\mu_\psi(f)+a^2v_\psi .
\]
Minimizing this one-dimensional quadratic gives
\[
    a^\star=\frac{\mu_\psi(f)}{v_\psi},
    \qquad
    \E\left[(Y-f(X)-a^\star \psi(X))^2\right]
    =
    \sigma_r^2(f)-\frac{\mu_\psi(f)^2}{v_\psi}.
\]
Thus the population fraction of residual mean-square removable by the best
single correction along \(\psi\) is
\begin{equation}\label{eq:population-effect-size}
e(\psi;f)
\coloneqq
\frac{\mu_\psi(f)^2}{\sigma_r^2(f)\,v_\psi}.
\end{equation}
This is the population target of the empirical effect size. It measures the
fraction of residual mean-square removable by the best one-direction correction
along \(\psi\), whereas \(t_n(\psi)\) measures how reliably the corresponding
moment differs from zero.

On the diagnostic set, we estimate $\sigma_r^2(f)$ by
\[
    \hat\sigma_r^2
    \coloneqq
    \frac1n\sum_{i=1}^n \hat r_i^2 .
\]
For probes normalized so that
\(n^{-1}\sum_i\psi(x_i)^2=1\), we estimate
\eqref{eq:population-effect-size} by
\begin{equation}\label{eq:effect-size}
e_n(\psi)
\coloneqq
\frac{\hat\mu_\psi^2}{\hat\sigma_r^2}
=
\frac{T_n(\psi)^2}{n\,\hat\sigma_r^2},
\qquad
\hat\mu_\psi
\coloneqq
\frac1n\sum_{i=1}^n \hat r_i\psi(x_i).
\end{equation}
If the probe is not normalized, the denominator should also include the
empirical factor \(n^{-1}\sum_i\psi(x_i)^2\). Unlike \(t_n(\psi)\),
\(e_n(\psi)\) does not grow merely because the diagnostic set is larger. It is therefore the quantity used to decide whether the removable residual
mean-square associated with \(\psi\) is large enough to justify deterministic
refinement. The threshold
\(\varepsilon\) should be a task-level tolerance fixed independently of \(n\).

\textcolor{black}{The diagnostic is run over a family of probes, and the reported quantity is the largest effect size over that family. Since \(e_n(\psi)\) measures the residual mean-square removable along \(\psi\), the conclusion does not depend on the particular probe family chosen, only on whether the family spans the directions along which residual structure may lie. Appropriate probe families include truncations of a complete basis, such as polynomials, Fourier modes, or random features. A family that probes more directions outside the model's fitting basis makes the diagnostic more sensitive, since residual structure lying entirely outside the probed span cannot be detected.}

\subsubsection{Residual mean-square and the aleatoric floor}

The residual mean-square plays a separate role. The effect size asks whether
a residual trend is large enough to remove. The residual mean-square asks
whether the error left after such trends have been removed is acceptable for
the scientific task.

From \eqref{eq:det-decomp}, the population residual risk of \(\hat f_n\) is
\[
\E\bigl[(Y-\hat f_n(X))^2\bigr]
=
\E\bigl[(m(X)-\hat f_n(X))^2\bigr]
+
\E[\Var(Y\mid X)].
\]
The first term is removable deterministic error: it is the gap between the
current surrogate and the conditional mean. The second term is the aleatoric
floor left by the chosen input \(X\). Thus \(\hat\sigma_r^2\) estimates this
floor from above. If enriched probe families find no practically important
residual structure and the residual mean-square stabilizes under further
deterministic refinement, the approximation gap has been driven below the
resolution of the diagnostic. At that point, \(\hat\sigma_r^2\) can be read as
an empirical estimate of the remaining conditional variability.

This step is what turns the residual test into a barrier diagnostic. It says
that, for the chosen descriptor \(X\), the fitted squared-loss surrogate has exhausted the detectable deterministic structure. The remaining decision is no longer whether to add capacity, but whether the scientific task can tolerate the estimated floor. In examples such as the Lorenz--96 closure of Section~\ref{sec:experiments-l96}, this is how the procedure locates an aleatoric floor that is not available in closed form.

\begin{remark}[Explained-variance ceiling]
For scalar \(Y\) with \(\Var(Y)>0\), the aleatoric floor can be expressed as a
scale-free ceiling on the coefficient of determination,
\[
R^2(f)
=
1-\frac{\E[(Y-f(X))^2]}{\Var(Y)} .
\]
\textcolor{black}{Combining the squared-loss decomposition \eqref{eq:det-decomp} with the law of
total variance \(\Var(Y)=\Var(m(X))+\E[\Var(Y\mid X)]\), any square-integrable
deterministic predictor \(f(X)\) satisfies
\[
R^2(f)\le R^2(m)
=
1-\frac{\E[\Var(Y\mid X)]}{\Var(Y)}
=
\frac{\Var(m(X))}{\Var(Y)},
\qquad m(X)=\E[Y\mid X],
\]
with equality if and only if \(f(X)=m(X)\) almost surely. Thus \(R^2(m)\) is the
largest coefficient of determination attainable by any deterministic
squared-loss predictor using only the information in \(X\): the ratio \(R^2(m)\)
is the fraction of total variance explained by the conditional mean}, while
\[
1-R^2(m)
=
\frac{\E[\Var(Y\mid X)]}{\Var(Y)}
\]
is the normalized aleatoric floor.

This ceiling can be interpreted together with the residual diagnostics.
An \(R^2\) plateau below one is evidence of the ceiling only after the probes
find no practically important residual structure. Before that point, the plateau
may still reflect a remaining approximation gap to \(m\). Once the gap has been
removed, the shortfall \(1-R^2(m)\) is the fraction of variance left unresolved
by the chosen input \(X\).
\end{remark}

\subsubsection{Decision rule}

The diagnostic can now be stated as a finite-data decision procedure.

\begin{procedure}{Barrier diagnostic for squared-loss surrogates}
\label{proc:residual-diagnostic}
\item \emph{Inputs.} Start from a fitted deterministic surrogate \(\hat f_n\),
an independent diagnostic set, a nested schedule of probe families
\(\Psi_1\subset\Psi_2\subset\cdots\), and a task-level tolerance
\(\varepsilon\) for the fraction of residual variance worth removing.

\item \emph{Residual probes.} For each probe \(\psi\), compute the
significance statistic \(t_n(\psi)\) and the effect size \(e_n(\psi)\). In
feature-based models the probes can be chosen outside the fitted feature
span, for instance polynomial degrees higher than the fitted degree. For
black-box surrogates, they should be pre-specified diagnostic transforms that
were not imposed as training constraints or used for model selection.

\item \emph{Stop rule.} Declare the conditional-mean barrier reached when no
diagnostic probe satisfies both \(|t_n(\psi)|>c_\alpha\) and
\(e_n(\psi)>\varepsilon\), and when this conclusion remains stable as the
probe family is enriched. In practice, the effect-size condition
\(\max_{\psi\in\Psi_j} e_n(\psi)\le\varepsilon\) is the primary stopping
criterion, while significance identifies reliable residual directions. Once
the residual mean-square \(\hat\sigma_r^2\) also stops decreasing beyond
sampling variation, read it as the empirical aleatoric floor for the chosen
input \(X\).

\item \emph{Output.} Combine residual structure and residual magnitude using
Table~\ref{tab:regimes}. The output is a modeling regime: refine the
deterministic surrogate, stop, or change the loss to score the feature of
\(P(Y\mid X)\) required by the task.
\end{procedure}

\begin{table}[t]
\centering
\caption{The diagnostic decision rule. Rows: does the orthogonality test
detect reducible out-of-basis structure? Columns: is the residual mean-square
$\hat\sigma_r^2$ within the downstream task's tolerance? Each cell gives the
regime and the action it licenses.}
\label{tab:regimes}
\small
\renewcommand{\arraystretch}{1.4}
\begin{tabularx}{\linewidth}{@{}
>{\raggedright\arraybackslash}p{0.20\linewidth}
>{\raggedright\arraybackslash}X
>{\raggedright\arraybackslash}X
@{}}
& \textbf{Residual within tolerance}\newline
  {\normalfont\small ($\hat\sigma_r^2 \le$ task tolerance)}
& \textbf{Residual exceeds tolerance}\newline
  {\normalfont\small ($\hat\sigma_r^2 >$ task tolerance)} \\
  \midrule
\textbf{Structure detected}\newline
{\normalfont\small (some $\psi$: $t_n$ significant, $e_n>\varepsilon$)}
& \emph{Regime~I$'$ (refine).} A reducible trend remains but the error is
already acceptable. Extracting it is optional; the barrier is not the
binding constraint.
& \emph{Regime~I (underfitting).} Reducible structure and intolerable error.
Act on the flagged directions: enlarge the class, change the basis, improve
optimization, or collect more data. \\
\textbf{No structure detected}\newline
{\normalfont\small (all $\psi$: $e_n\le\varepsilon$)}
& \emph{Regime~II (adequate).} $\hat f_n\approx m$ and the floor is
acceptable. Stop; deterministic squared-loss modeling suffices.
& \emph{Regime~III (the barrier).} $\hat f_n\approx m$ yet the residual is
intolerable: the error is dominated by $\E[\Var(Y\mid X)]$. Enlarging the
deterministic class cannot help---change the loss, not the model. \\
\bottomrule
\end{tabularx}
\renewcommand{\arraystretch}{1}
\end{table}

This procedure can lead to a conclusion for modeling decision: whether to keep improving a deterministic squared-loss surrogate or to change the target. Such a decision should not be based on a single non-rejection, because failure to detect residual structure for one probe family only means that no practically important structure was found at that diagnostic resolution. The conclusion should instead remain stable as the probe family is enriched. Two additional safeguards are needed to make the diagnostics robust. First, the residual moments should be
evaluated on a diagnostic set independent of the data used to fit the
surrogate. Otherwise, training residuals may look artificially small or
orthogonal because of the fitting procedure itself. Second, the probe functions
should test directions that were not already forced to vanish by the training
objective. For feature-based least squares, this means out-of-basis
directions; for black-box surrogates, it means probe functions of \(X\) that
were not imposed through the training loss, constraints, or model-selection
criterion.

\section{After the barrier: choosing distributional objectives}
\label{sec:distributional}

Regime~III in Table~\ref{tab:regimes} identifies a specific modeling
situation: a deterministic squared-loss surrogate has no practically important
residual structure left in the chosen input \(X\), but the remaining variability
is still too large for the scientific task. The question is then no longer how
to refine the point predictor, but which feature of the conditional law
\(P(Y\mid X)\) the downstream computation needs.

\subsection{Randomness under squared loss is not enough}
\label{subsec:varsupp}

A natural reaction to the conditional-mean barrier is to make the surrogate
stochastic. Let
\[
    G(X,Z)
\]
be a stochastic predictor, where \(Z\sim p_Z\) is an auxiliary latent variable
sampled independently of \(Y\) conditional on \(X\). For fixed \(X=x\), the
randomness in \(Z\) induces a model distribution for the output. This can
represent variability only if the loss scores the right kind of variability.
Paired squared loss does not.

With \(m(X)=\E[Y\mid X]\), {\color{black}a bias--variance decomposition of the conditional risk, followed by taking expectations over \(X\), gives}
\[
\E\bigl[(Y-G(X,Z))^2\bigr]
=
\E[\Var(Y\mid X)]
+
\E\bigl[(m(X)-\E_Z[G(X,Z)\mid X])^2\bigr]
+
\E[\Var_Z(G(X,Z)\mid X)].
\]
The first term is the aleatoric floor associated with the chosen input \(X\):
it is the average conditional variance under \(P(Y\mid X)\), and does not
depend on the fitted model. The last two terms are non-negative and are
minimized, in an expressive class, by
\[
    \E_Z[G(X,Z)\mid X]=m(X),
    \qquad
    \Var_Z(G(X,Z)\mid X)=0
    \quad \text{a.s.}
\]
Thus latent randomness alone does not overcome the barrier when the objective
remains paired squared loss. The loss suppresses model variance and collapses
the stochastic predictor back to the conditional mean. After Regime~III has
been diagnosed, the repair must change the scored quantity, not merely the
sampler.

\subsection{Likelihood as a minimal distributional objective}
\label{subsec:nll}

NLL is the most direct example of such a change. Suppose the model specifies a conditional density \(q_\theta(y\mid x)\). The population NLL is
\[
\mathcal L_{\mathrm{NLL}}(\theta)
=
\E[-\log q_\theta(Y\mid X)].
\]
When the true conditional law admits a density \(p(\cdot\mid x)\) with respect
to the same reference measure,
\[
\E[-\log q_\theta(Y\mid X)]
=
\E_X\!\left[
\KL\bigl(p(\cdot\mid X)\,\Vert\,q_\theta(\cdot\mid X)\bigr)
\right]
+
\text{constant}.
\]
Thus NLL changes the population target from the conditional mean to the Kullback–Leibler (KL) projection of the conditional law within the chosen density family.

A heteroscedastic Gaussian model illustrates the minimal repair. For scalar
outputs, let
\[
q_{\theta_\mu,\theta_s}(y\mid x)
=
\mathcal N\bigl(y;\mu_{\theta_\mu}(x),s_{\theta_s}(x)\bigr),
\qquad
s_{\theta_s}(x)>0,
\]
where \(s_{\theta_s}(x)\) denotes the conditional variance. The population NLL,
up to an additive constant, is
\[
\E\left[
\frac12\log s_{\theta_s}(X)
+
\frac{(Y-\mu_{\theta_\mu}(X))^2}{2s_{\theta_s}(X)}
\right].
\]
\textcolor{black}{This objective separates pointwise in \(x\); minimizing the unrestricted problem over measurable \(\mu(x)\) and \(s(x)>0\) for each fixed \(x\) gives, when \(\Var(Y\mid X=x)>0\),
\[
\mu^\star(x)=\E[Y\mid X=x],
\qquad
s^\star(x)=\Var(Y\mid X=x),
\]
with the same conclusion in the degenerate limit \(s^\star(x)\downarrow 0\) when the conditional variance vanishes.}
Thus Gaussian NLL still targets
the conditional mean, but it also scores the conditional scale. In a
parameterized model, \(\mu_{\theta_\mu}\) and \(s_{\theta_s}\) approximate
these population targets within the chosen function classes. This is a
minimal beyond-barrier objective when the scientific task needs calibrated
spread rather than only a mean response. It does not represent multimodality
unless the density family is enriched, for example by mixtures, flows, or
other conditional density models.

\subsection{A post-diagnostic map of objectives}
\label{subsec:objective-map}

The broader lesson is that there is no universally correct distributional
repair. Once the diagnostic has separated removable deterministic structure
from residual conditional variability, the next loss should be chosen
according to the scientific quantity needed downstream. If only the mean
matters, squared loss remains appropriate. If uncertainty intervals or
likelihoods matter, a proper probabilistic scoring rule is natural. If the
downstream computation is sensitive to spectra, fluxes, moments, or structure
functions, then those observables may be better targets than a full density.
If plausible high-dimensional samples are required, implicit or score-based
samplers may be the relevant objects.

Table~\ref{tab:map} summarizes this post-diagnostic choice. The objectives in
the table are standard; the point here is how they enter the decision pipeline:
they are invoked only after the residual diagnostics indicate that a
squared-loss point predictor is no longer aligned with the task.

\begin{table}[t]
\centering
\caption{Post-diagnostic objective selection. The loss should score the
feature of \(P(Y\mid X)\) required by the downstream scientific computation.}
\label{tab:map}
\small
\renewcommand{\arraystretch}{1.3}
\begin{tabularx}{\linewidth}{@{}
>{\raggedright\arraybackslash}p{0.28\linewidth}
>{\raggedright\arraybackslash}X
>{\raggedright\arraybackslash}X
@{}}
Scientific need & Feature of \(P(Y\mid X)\) to score & Representative objectives or models \\
\midrule
Only the mean response matters
& Conditional mean \(\E[Y\mid X]\)
& Deterministic surrogate trained by squared loss \\

\textcolor{black}{Calibrated spread, or surrogate stochastic differential equations}
& Conditional density or probabilistic forecast
& \textcolor{black}{NLL~\citep{dietrich2023learning,zhu2024dyngma}; mixture densities~\citep{zhu2024dyngma};
normalizing flows~\citep{papamakarios2021normalizing,guo2022normalizing,yang2024pseudoreversible}} \\

Task-relevant scientific statistics
& Selected moments, fluxes, spectra, structure functions, or other observables
& Moment or observable matching
\citep{hansen1982large,cleary2021calibrate,qi2023data} \\

Latent-variable density approximation
& Conditional likelihood through a latent representation
& Conditional variational autoencoders and ELBO objectives
\citep{kingma2014auto,rezende2014stochastic,gundersen2021semi} \\

Plausible high-dimensional samples
& Conditional sample distribution, possibly through a learned discrepancy
& Adversarial objectives and conditional GANs
\citep{goodfellow2014generative,mirza2014conditional,yang2020physics};
diffusion or score-based models
\citep{song2021score,ho2020denoising,vincent2011connection,liu2025training} \\
\bottomrule
\end{tabularx}
\renewcommand{\arraystretch}{1}
\end{table}

The table is meant as a scope control, not as a survey of generative modeling.
Its role is to make the post-diagnostic decision explicit: after the
conditional-mean barrier has been identified, one should choose a loss that
scores the feature of \(P(Y\mid X)\) required by the scientific computation.

\section{Numerical studies}\label{sec:experiments}
The numerical studies serve two roles in the diagnostic pipeline. The
controlled two-branch law validates the diagnostic when all population
quantities are known, while the two-scale Lorenz--96 closure problem shows the
same decision process in a scientific-computing setting where the aleatoric
floor must be inferred from data. The goal is not to benchmark closure methods,
but to show how the diagnostic separates deterministic underfitting from the
conditional-mean barrier and how a minimal post-diagnostic change of loss
affects downstream statistics. All scripts run on CPU by default, with fixed
random seeds and single-threaded execution to reduce backend-dependent
variation. We use polynomial regressors throughout: Legendre polynomials in the one-dimensional controlled example and total-degree polynomials of standardized stencil variables in the Lorenz--96 closure. Code and data are available at \url{https://github.com/junfeng-chen/conditional_mean_barrier}.

\subsection{Controlled validation on a two-branch law}\label{subsec:exp-twobranch}

We first return to the two-branch construction of
Example~\ref{ex:two-branch}, now with all population quantities known in
closed form. Let
\[
X\sim \mathrm{Unif}[-1,1],\qquad
B\in\{-1,1\},\qquad
\mathbb P(B=1)=\mathbb P(B=-1)=\frac12,
\]
and define
\begin{equation}\label{eq:exp-two-branch-law}
Y=m(X)+B\,a(X),\qquad
m(x)=\sin(2\pi x),\qquad
a(x)=0.5+0.3\cos(2\pi x).
\end{equation}
Then
\[
\E[Y\mid X=x]=m(x),
\qquad
\Var(Y\mid X=x)=a(x)^2.
\]
For this law,
\[
\Var(m(X))=\frac12,
\qquad
\E[\Var(Y\mid X)]=\E[a(X)^2]=0.295,
\]
and therefore
\begin{equation}\label{eq:exp-r2-ceiling}
\Var(Y)=0.795,\qquad
R^2(m)=\frac{\Var(m(X))}{\Var(Y)}
      =\frac{0.5}{0.795}\approx 0.629 .
\end{equation}
Thus, even a perfect squared-loss predictor cannot attain \(R^2=1\):
roughly \(37\%\) of the variance in \(Y\) is irreducible conditional
variability.

We generated \(800\) training samples, \(2,000\) independent diagnostic
samples, and \(5,000\) independent test samples from \eqref{eq:exp-two-branch-law}. The diagnostic set is chosen larger than the training set so that residual moments are estimated with visibly smaller sampling noise, while the test set is used only to estimate the \(R^2\) curve. Deterministic predictors were fitted by least squares in the Legendre basis
\[
\hat f_p(x)=\sum_{j=0}^{p}\hat\beta_j P_j(x),
\]
where \(P_j\) is the \(j\)-th Legendre polynomial on \([-1,1]\). We use
\(p=2\) as a deliberately underfit model and \(p=9\) as a high-capacity
representative. The whole degree path \(p=0,\ldots,30\) is also computed
to visualize the explained-variance ceiling.

\begin{figure}[t]
\centering

\subfloat[Squared-loss regression learns the conditional mean.\label{fig:twobranch-fits}]{%
\begin{minipage}[b]{0.47\linewidth}
\centering
\includegraphics[width=\linewidth]{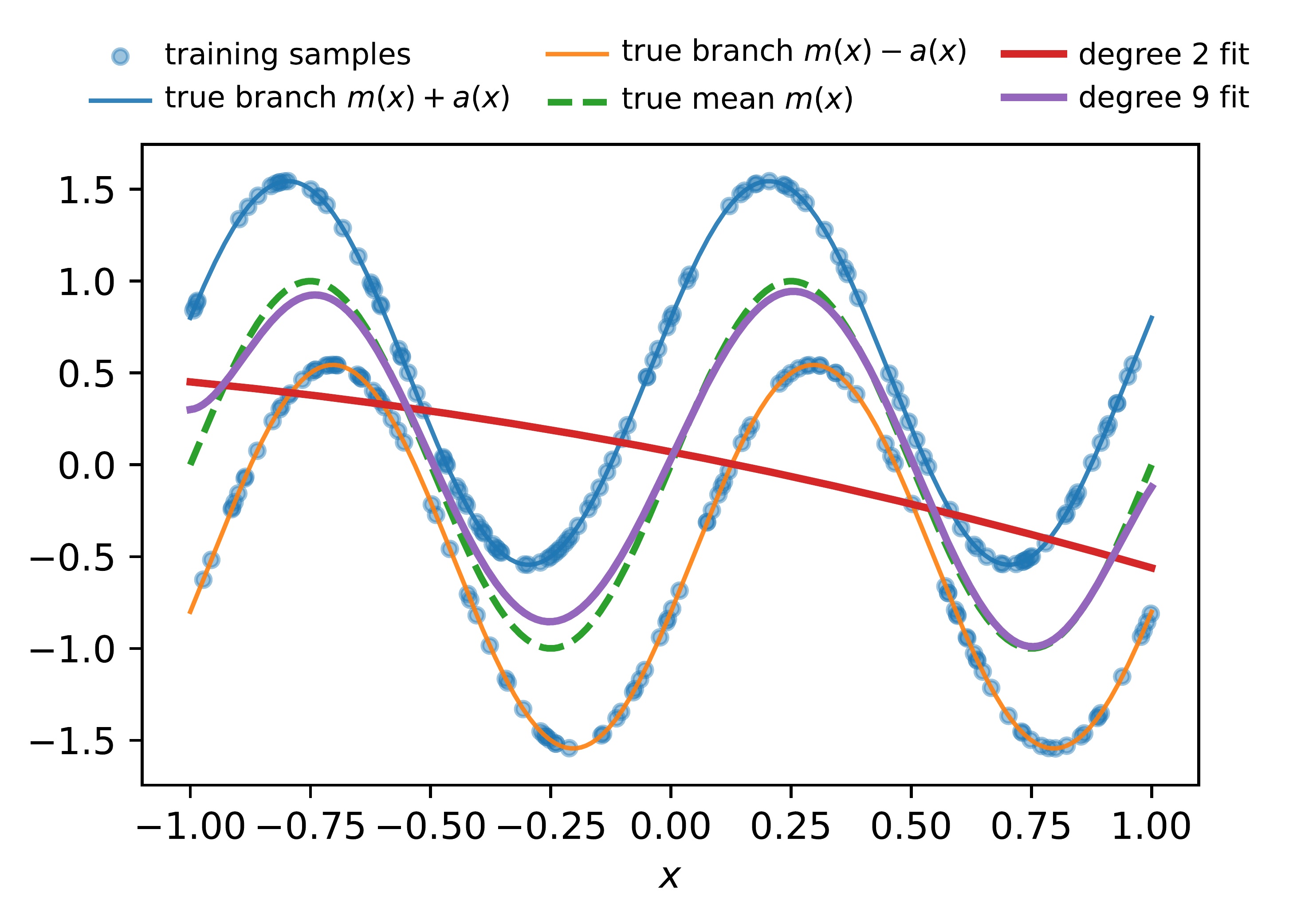}
\end{minipage}
}
\hfill
\subfloat[\(R^2\) saturates at the explained-variance ceiling.\label{fig:twobranch-r2}]{%
\begin{minipage}[b]{0.46\linewidth}
\centering
\includegraphics[width=\linewidth]{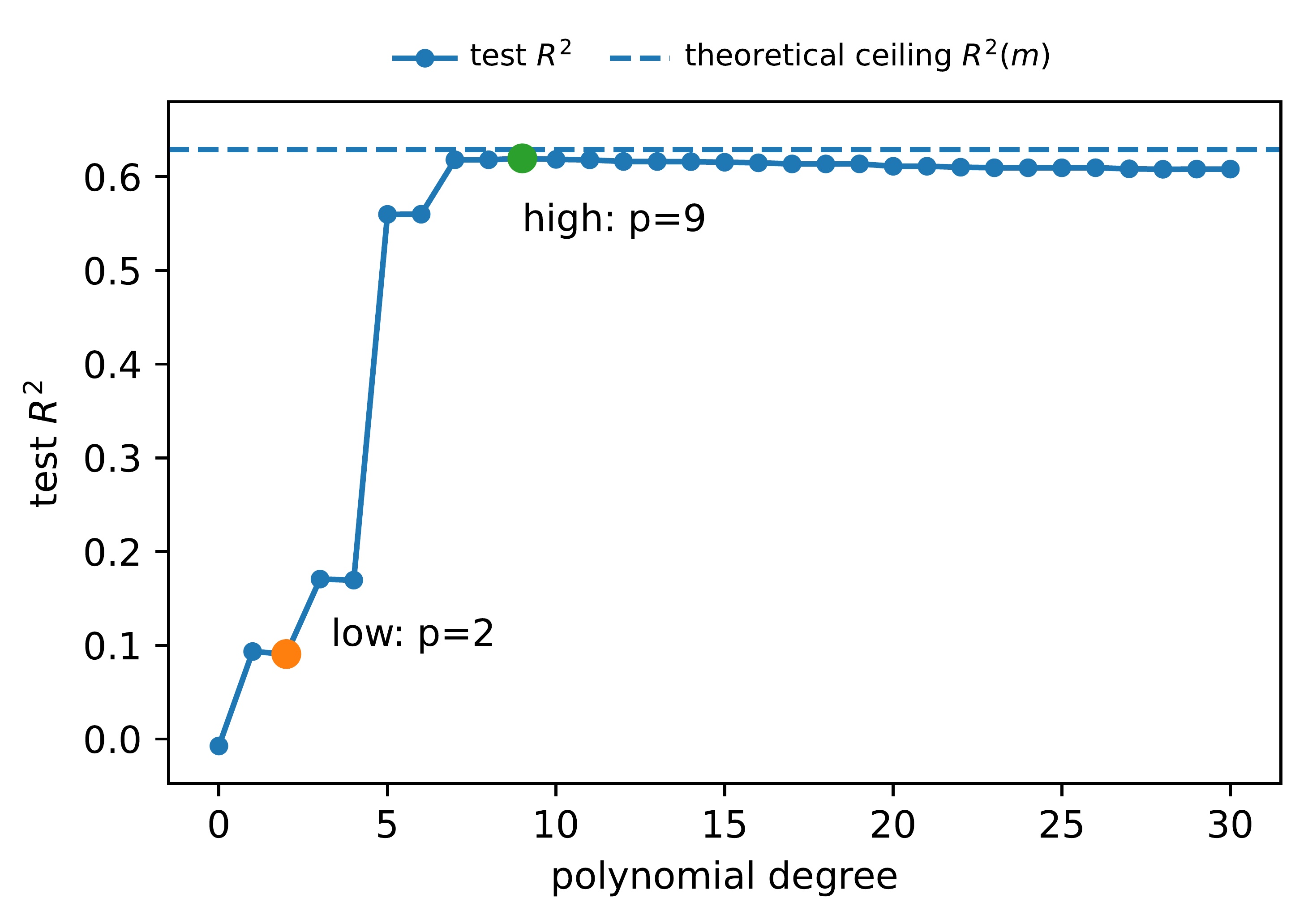}
\end{minipage}
}

\caption{A controlled two-branch experiment. (a) Samples from
\eqref{eq:exp-two-branch-law}, the two branches \(m(x)\pm a(x)\), the true
conditional mean \(m(x)\), and two deterministic least-squares fits. (b) Test
\(R^2\) versus polynomial degree, saturating at the explained-variance ceiling
\eqref{eq:exp-r2-ceiling}.}
\label{fig:twobranch-main}
\end{figure}

Figure~\ref{fig:twobranch-main}(a) shows the sampled two-branch law, the
true conditional mean \(m\), and the two least-squares fits. The degree-2
model misses the oscillatory conditional mean and leaves substantial
deterministic structure in the residual. By contrast, the degree-9 model
tracks \(m\) closely. Notice that the fitted curve lies between the two
branches: it is the squared-loss-optimal deterministic summary, not a
typical realization of \(Y\mid X=x\).

Figure~\ref{fig:twobranch-main}(b) plots the test \(R^2\) as the polynomial
degree increases. The curve rises quickly and then saturates near the
theoretical ceiling \eqref{eq:exp-r2-ceiling}. For \(p=9\), the test error
is approximately
\[
\mathrm{MSE}_{\rm test}(\hat f_9)\approx 0.302,
\qquad
R^2_{\rm test}(\hat f_9)\approx 0.620,
\]
close to the Bayes risk \(0.295\) and the ceiling \(0.629\). Increasing
capacity beyond this point does not move the deterministic predictor toward
\(R^2=1\), because the remaining error is dominated by
\(\E[\Var(Y\mid X)]\).

The residual-feature diagnostic gives the complementary view, pairing the two
statistics of Procedure~\ref{proc:residual-diagnostic}. On the \(n=2000\)
diagnostic samples we evaluate the significance \(t_n(\psi)\) and the effect
size \(e_n(\psi)\) (see equations \eqref{eq:significance} and \eqref{eq:effect-size}), taking \(\psi\) to be the orthonormalized Legendre
probes \(\sqrt{2k+1}\,P_k\); the factor \(\sqrt{2k+1}\) gives unit variance
under \(X\sim\mathrm{Unif}[-1,1]\), so \(e_n\) reads directly as the fraction
of residual variance a correction along \(P_k\) would remove. We fix the task
tolerance at \(\varepsilon=0.01\).

\begin{figure}[t]
\centering

\subfloat[Residual diagnostics: significance.\label{fig:twobranch-significance}]{%
\begin{minipage}[b]{0.465\linewidth}
\centering
\includegraphics[width=\linewidth]{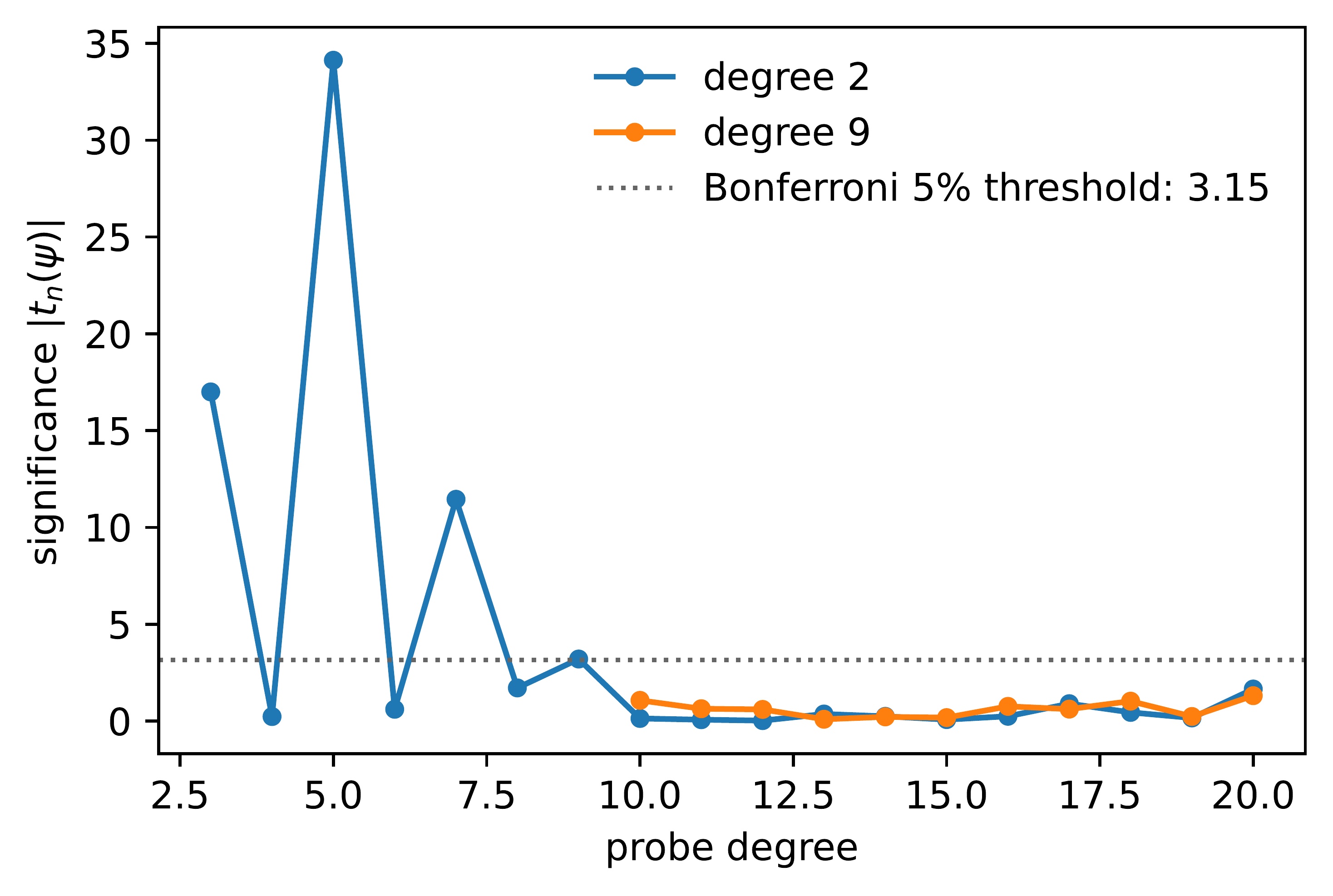}
\end{minipage}
}
\hfill
\subfloat[Residual diagnostics: effect size.\label{fig:twobranch-effectsize}]{%
\begin{minipage}[b]{0.47\linewidth}
\centering
\includegraphics[width=\linewidth]{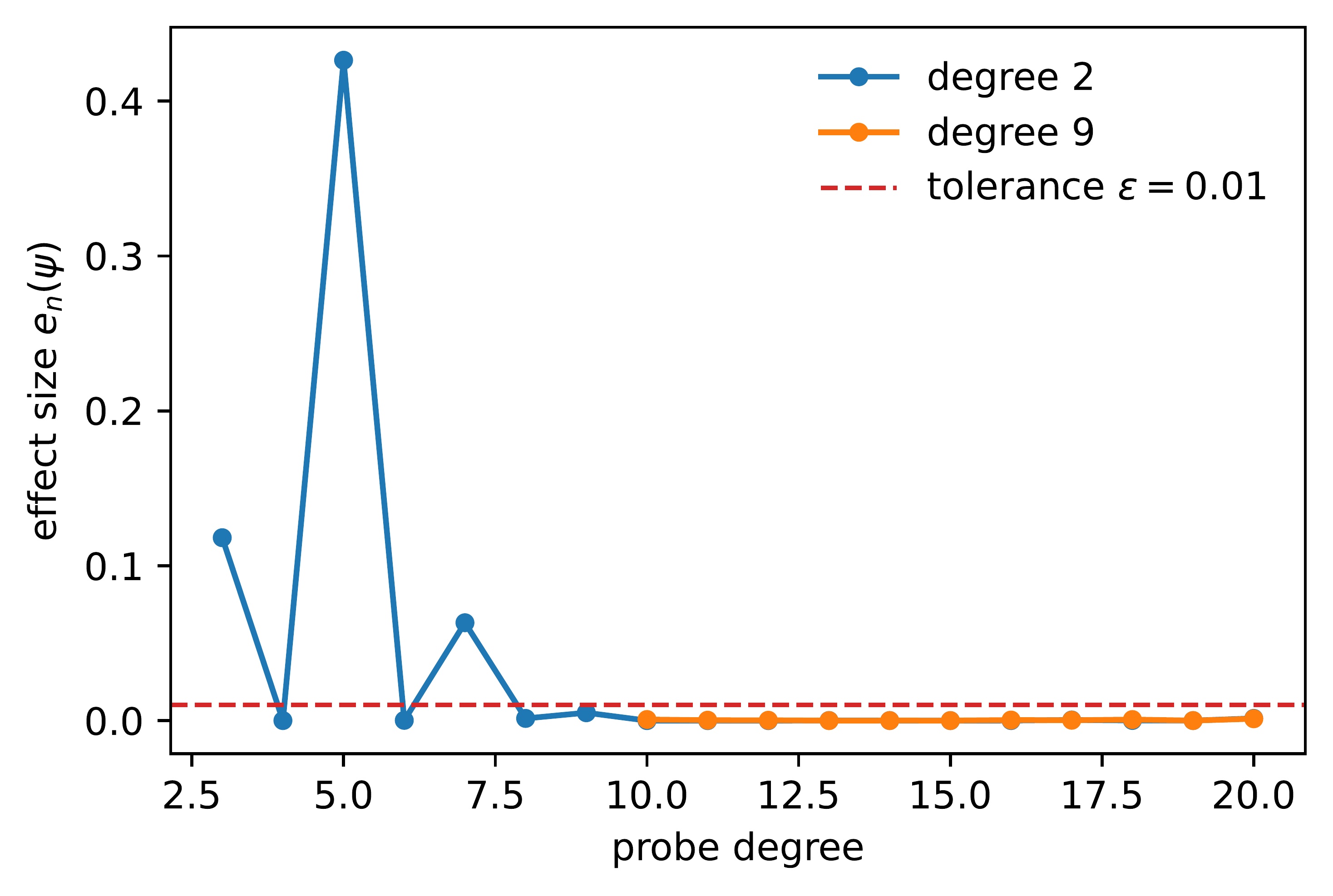}
\end{minipage}
}

\caption{Residual-feature diagnostics for the two-branch example, for the
degree-2 (underfit) and degree-9 (high-capacity) least-squares fits, evaluated
on out-of-basis Legendre probes. (a): significance \(|t_n(P_k)|\), with the
Bonferroni-corrected \(5\%\) line \(|t_n|=3.15\). (b): effect size \(e_n(P_k)\), the fraction of residual variance a correction along \(P_k\) would remove, with
the task tolerance \(\varepsilon=0.01\). The degree-2 residual is both significant and large (Regime~I); the degree-9 residual is neither (Regime~III) and its mean-square equals the aleatoric floor.}
\label{fig:twobranch-diagnostics}
\end{figure}

Figure~\ref{fig:twobranch-diagnostics} reports both statistics for the
out-of-basis probes. For the degree-2 fit the residual carries structure
that is both significant and practically large: the worst direction \(P_5\)
reaches \(|t_n|\approx 34\) and \(e_n\approx 0.43\), so a single Legendre
correction would remove about \(43\%\) of the residual variance, far above
\(\varepsilon\). Together with its large prediction error this places the
degree-2 model in Regime~I (deterministic underfitting, Table~\ref{tab:regimes}):
the flagged directions are exactly the missing oscillatory structure, and
enlarging the basis is the right remedy. For the degree-9 fit no out-of-basis
probe is either significant or large---maximum \(|t_n|\approx 1.3\) and
maximum \(e_n\approx 1\times10^{-3}\), well below \(\varepsilon\)---so the
orthogonality diagnostic is satisfied. Its residual mean-square is not small,
however: on the diagnostic set \(\hat\sigma_r^2\approx 0.296\), essentially
the Bayes floor \(\E[\Var(Y\mid X)]=0.295\). The degree-9 fit therefore sits
at the conditional-mean barrier (Regime~III): it has reached \(m\), the
diagnostic has located the aleatoric floor from data, and the remaining error
is irreducible by any squared-loss method. The same prediction-error level
thus carries two different verdicts---learnable deterministic structure for
\(p=2\), the barrier for \(p=9\)---and it is the pairing of significance with
effect size, not either alone, that separates them.

\subsection{Two-scale Lorenz--96 closure modeling}
\label{sec:experiments-l96}

We now turn to the main closure-modeling study, where the population quantities
are not available in closed form. The purpose is to show how the diagnostic
framework can be used when the conditional-mean barrier must be inferred from
data, and how that diagnosis predicts a downstream failure mode of a
deterministic mean closure.

We use the two-scale Lorenz--96 system \citep{lorenz1996predictability,wilks2005effects}
\begin{align}
\dot X_k
&=
(X_{k+1}-X_{k-2})X_{k-1}-X_k+F
-\frac{hc}{b}\sum_{j=1}^{J}Y_{j,k},
\label{eq:l96-slow}
\\
\dot Y_{j,k}
&=
-cbY_{j+1,k}(Y_{j+2,k}-Y_{j-1,k})
-cY_{j,k}
+\frac{hc}{b}X_k,
\label{eq:l96-fast}
\end{align}
with periodic indexing in both \(k\) and \(j\). The slow variables
\(\{X_k\}_{k=1}^{K}\) are resolved, while the fast variables
\(\{Y_{j,k}\}\) are treated as unresolved. The unresolved forcing acting on
the \(k\)-th slow variable is
\begin{equation}
C_{t,k}
=
-\frac{hc}{b}\sum_{j=1}^{J}Y_{j,k}(t).
\label{eq:l96-closure-target}
\end{equation}
The closure problem is to approximate \(C_{t,k}\) from resolved information.
In this experiment we choose the local stencil
\begin{equation}
S_{t,k}
=
\bigl(X_{k-1}(t),X_k(t),X_{k+1}(t)\bigr)
\label{eq:l96-local-stencil}
\end{equation}
as the descriptor and fit deterministic maps
\[
\hat f_p(S_{t,k})\approx C_{t,k}.
\]
This choice is intentionally local and interpretable. A wider stencil or time
delay embeddings would define a different conditioning variable and hence a
different conditional law; here we ask what can be learned from the fixed
local stencil \eqref{eq:l96-local-stencil}.

We use \(K=J=8\), \(F=10\), \(h=2\), and \(c=b=10\). The full two-scale
system is integrated with time step \(\Delta t=5\times 10^{-3}\). After a
burn-in period, the trajectory is downsampled to reduce short-time
correlation and then split into time blocks. The retained data contain
\(300\) snapshots, hence \(300K=2400\) local stencil--closure pairs. The
first \(180\) snapshots are used for fitting, the next \(60\) for diagnostics,
and the last \(60\) for held-out one-step evaluation.

\begin{figure}[t]
    \centering
    \includegraphics[width=\textwidth]{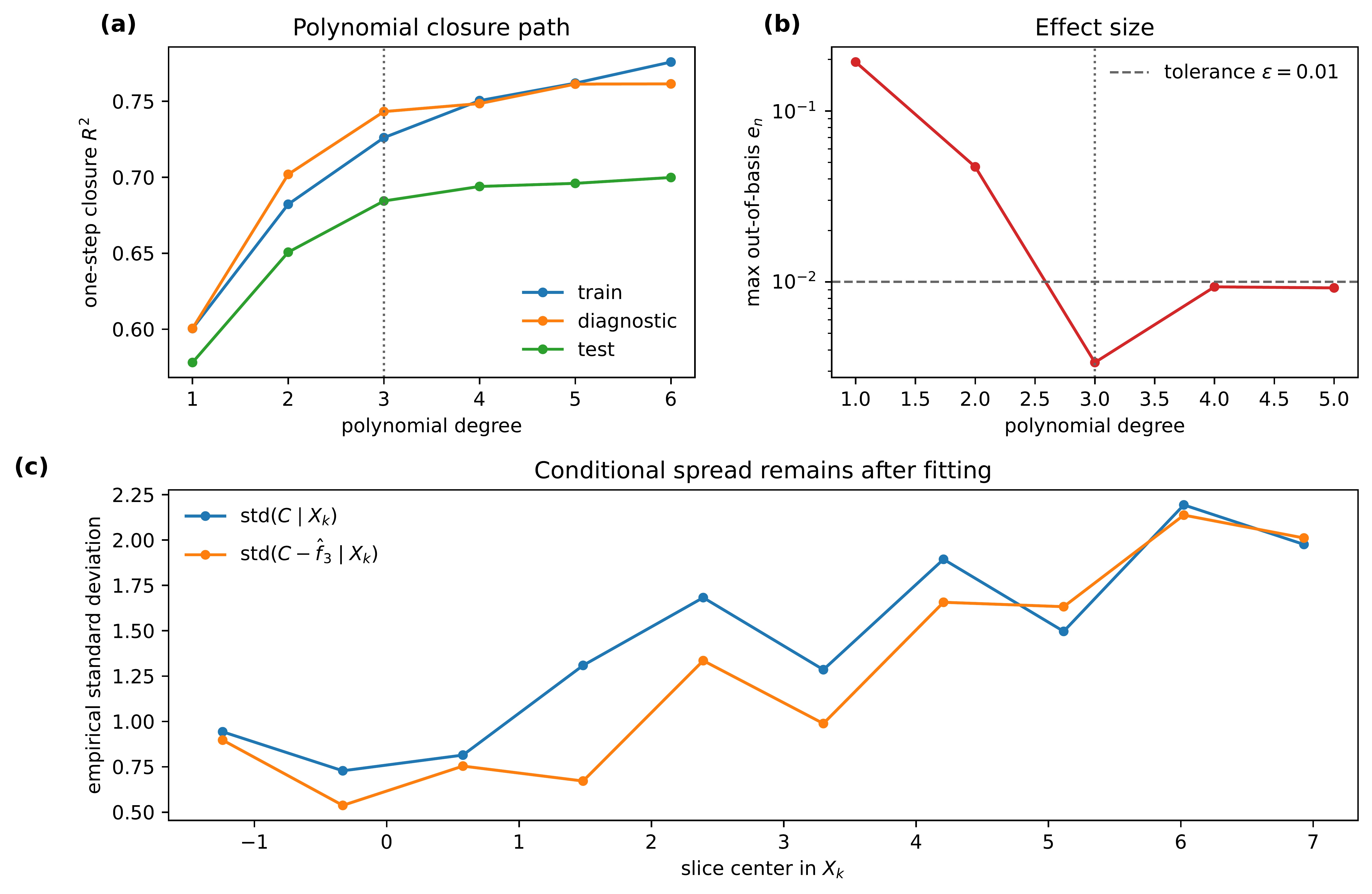}
    \caption{
    Diagnostic results for the two-scale Lorenz--96 closure.
    (a) One-step closure \(R^2\) (train, diagnostic, test) versus polynomial
    degree, saturating by \(p=3\) (dotted line: selected closure).
    (b) Primary diagnostic: the maximum out-of-basis effect size
    \(\max_\psi e_n(\psi)\) versus degree (log scale), with the task tolerance
    \(\varepsilon=0.01\); the maximum drops below \(\varepsilon\) at \(p=3\),
    locating the conditional-mean barrier.
    (c) Conditional spread after mean fitting: within ten bins of the central
    stencil component \(X_k\) (neighbors \(X_{k\pm1}\) free), the standard
    deviation of the true forcing \(\operatorname{std}(C\mid X_k)\) and of the
    post-closure residual \(\operatorname{std}(C-\hat f_3\mid X_k)\).}
    \label{fig:l96-diagnostics}
\end{figure}

For each degree \(p\), we fit a total-degree polynomial closure
\(\hat f_p\) on standardized stencil variables by least squares.
Figure~\ref{fig:l96-diagnostics}(a) shows the one-step closure accuracy: the
train, diagnostic, and test \(R^2\) all rise rapidly from \(p=1\) to \(p=3\)
and then saturate, with the test \(R^2\) remaining slightly below the
diagnostic \(R^2\) throughout.

We select the representative closure by Procedure~\ref{proc:residual-diagnostic},
not by maximizing one-step \(R^2\) alone.
The probe family is the monomials of the standardized stencil up to total
degree~6; a probe is \emph{out-of-basis} for a degree-\(p\) fit if its total
degree exceeds \(p\). For each fit we compute the effect size \(e_n(\psi)\) of
\eqref{eq:effect-size} on the diagnostic snapshots---the fraction of residual variance a correction along \(\psi\) would remove. To remove the influence of the spatial correlation among the \(K\) stencils recorded at one time step, we report the snapshot-clustered results.
Figure~\ref{fig:l96-diagnostics}(b) plots \(\max_\psi e_n(\psi)\) over the
out-of-basis probes against the fitted degree, with the task tolerance fixed at \(\varepsilon=0.01\), the same value as in
Section~\ref{subsec:exp-twobranch}. \textcolor{black}{As the degree increases, the largest out-of-basis effect size falls sharply across \(p=1,2,3\) (\(\max_\psi e_n = 0.193,\,0.047,\,0.0034\)), crossing the tolerance \(\varepsilon\) at \(p=3\) and remaining below it at \(p=4,5\). Enriching the basis further removes no practically large structure, so the stop rule is satisfied at \(p=3\), which we adopt for the subsequent analysis.} For the selected model,
\[
R^2_{\mathrm{diag}}=0.743,
\qquad
R^2_{\mathrm{test}}=0.684 .
\]
Equivalently, the residual mean-square on the diagnostic set is
\(\hat\sigma_r^2\approx 1.82\) against a forcing variance of \(7.10\), so that
about \(26\%\) of the unresolved forcing is irreducible from this stencil.
Once the stop rule holds, this \(\hat\sigma_r^2\) is the empirical estimate of
the aleatoric floor \(\E[\Var(C\mid S)]\), which is not available in closed
form here (Section~\ref{subsec:diagnostics}). This is the central diagnostic
result for the case study: the $p=3$ closure sits at the conditional-mean
barrier (Regime~III of Table~\ref{tab:regimes}).

A significance test corroborates the selection. Because the \(K=8\) stencils
in a snapshot are spatially correlated, raw per-row significance would be
overstated; we therefore form a snapshot-clustered statistic, averaging the
row product \(r_{t,k}\psi(S_{t,k})\) within each snapshot to a snapshot moment
\(H_t = K^{-1}\sum_k r_{t,k}\psi(S_{t,k})\) and computing the one-sample
\(t\)-statistic \(t_{59}(\psi)=\sqrt{60}\,\bar H/\hat s_H\). The clustered
\(\max_\psi|t_{59}|\) over out-of-basis probes falls
\(7.96\to 4.84\to 1.43\) across \(p=1,2,3\), dropping below the
Bonferroni-corrected \(5\%\) threshold (\(|t|\approx 3.6\)) at \(p=3\), in
agreement with the effect size. Consistent with the supporting role
significance plays in Section~\ref{subsec:diagnostics}, we treat the effect
size as the primary criterion and the clustered test as a check.

Reaching the barrier does not mean the forcing is determined by the local
stencil: a large conditional spread can persist around the learned mean. To
visualize this, Figure~\ref{fig:l96-diagnostics}(c) partitions the diagnostic
pairs \((S_{t,k},C_{t,k})\) into ten bins by the value of the central stencil
component \(X_k\), and plots the standard deviation of the true forcing \(\operatorname{std}(C\mid X_k)\) and of the post-closure residual \(\operatorname{std}(C-\hat f_3\mid X_k)\) against the bin mean of \(X_k\). Across the range of the resolved state the two curves nearly coincide, so the deterministic closure removes little of the spread.

Finally, we examine what happens when the fitted deterministic closure is
embedded into an autonomous slow-only model,
\begin{equation}
\dot{\widetilde X}_k
=
(\widetilde X_{k+1}-\widetilde X_{k-2})\widetilde X_{k-1}
-\widetilde X_k+F
+
\hat f_3(\widetilde S_k),
\label{eq:l96-closed-system}
\end{equation}
where
\[
\widetilde S_k
=
(\widetilde X_{k-1},\widetilde X_k,\widetilde X_{k+1}).
\]
The rollout comparison is deliberately separate from the previously used dataset. We generate new long trajectories of both the full two-scale reference system and the closed slow-only system starting from the same initial condition, discard burn-in, and compare downstream statistics.

The one-point marginal variance of \(X_k\) is nearly preserved:
\[
\frac{\operatorname{Var}_{\mathrm{closed}}(X_k)}
{\operatorname{Var}_{\mathrm{ref}}(X_k)}
\approx 1.000.
\]
This is an important caution: the closed slow-only system is itself a nonlinear
dynamical system with its own attractor, so its one-point marginals can match
those of the reference even when its joint or time-aggregated statistics do
not. The more revealing statistic is the slow energy
\begin{equation}
E(t)=\frac{1}{2K}\sum_{k=1}^{K}X_k(t)^2.
\label{eq:l96-slow-energy}
\end{equation}
For this quantity the deterministic closure strongly suppresses fluctuations. Table~\ref{tab:l96-downstream-ratios} summarizes the contrast, and the slow-energy densities of Figure~\ref{fig:l96-stochastic-summary}(a) show the prediction much narrower than the reference, its variance reduced to \(5.5\%\) of the reference value, although the marginal variance of \(X_k\) is essentially unchanged.

\begin{table}[t]
    \centering
    \begin{tabular}{lcc}
    \toprule
    Statistic & Closed/reference ratio & Interpretation \\
    \midrule
    \(\operatorname{Var}(X_k)\) & \(1.000\) & one-point marginal variance preserved \\
    \(\operatorname{Var}(E)\) & \(0.055\) & collective energy fluctuations suppressed \\
    \bottomrule
    \end{tabular}
    \caption{
    Downstream statistics from independent long rollouts of the full
    two-scale reference system and the closed deterministic slow-only system.
    }
    \label{tab:l96-downstream-ratios}
\end{table}

The slow-energy trajectory of Figure~\ref{fig:l96-stochastic-summary}(b) shows the dynamical side of the same effect: the closed deterministic system fluctuates in a visibly smaller band, while the space--time windows of Figure \ref{fig:l96-spacetime-triptych} confirm that the closed system remains dynamically active. The issue is not collapse to a constant state but distortion of collective fluctuations. This illustrates the
practical form of the conditional-mean barrier in closure modeling: fitting the mean is not the same as modeling the conditional law.

\textcolor{black}{\begin{remark}
Sections~\ref{subsec:exp-twobranch} and~\ref{sec:experiments-l96} use
out-of-basis polynomials (Legendre and monomial, respectively) as probe
functions, since the regressors are themselves finite-degree polynomials. However, the diagnosis is
model-agnostic: for black-box neural surrogates, the probes may instead be
generic nonlinear or random features. The conclusion is also insensitive to the
threshold \(\varepsilon\) whenever the effect size drops sharply across the
barrier; here it falls from \(0.047\) at \(p=2\) to \(0.0034\) at \(p=3\), so any
intermediate tolerance selects \(p=3\). In practice, \(\varepsilon\) should be
set case by case according to the tolerance of the downstream task.
\end{remark}}

\subsection{A minimal post-diagnostic stochastic-scale intervention}
\label{sec:experiments-l96-stochastic}

Section~\ref{sec:experiments-l96} established that the nearly optimal deterministic closure suppresses collective slow-energy fluctuations. To cross the conditional-mean barrier, we now construct a minimal stochastic closure---keeping the deterministic mean closure fixed and learning only a
conditional scale by NLL.

Let \(\hat f_3(S)\) denote the selected deterministic least-squares closure
from Section~\ref{sec:experiments-l96}. We fit a heteroscedastic Gaussian
approximation
\[
C_{t,k}\mid S_{t,k}
\approx
\mathcal N\!\left(\hat f_3(S_{t,k}),\hat\sigma(S_{t,k})^2\right),
\]
where \(\hat f_3\) is frozen and only \(\hat\sigma\) is trained. Equivalently,
on the fixed residuals
\[
r_{t,k}=C_{t,k}-\hat f_3(S_{t,k}),
\]
we minimize the Gaussian NLL
\begin{equation}
\mathcal L_{\mathrm{scale}}
=
\frac1n\sum_i
\left[
\log \sigma_\theta(S_i)
+
\frac{r_i^2}{2\sigma_\theta(S_i)^2}
\right],
\label{eq:l96-scale-nll}
\end{equation}
up to the irrelevant constant \(\frac12\log(2\pi)\).
{\color{black}This is the per-snapshot Gaussian transition-density likelihood used to identify the drift and diffusion of effective stochastic differential equation (SDEs) from discrete observations~\citep{dietrich2023learning}, here applied to the closure residuals with the mean held fixed; more accurate transition-density approximations~\citep{zhu2024dyngma} could be substituted when needed.}
The log-scale is modeled
by a low-degree polynomial and lightly regularized. Freezing the mean is
important: jointly fitting a heteroscedastic likelihood also reweights the
mean, which can change the closed-loop drift even when one-step scores remain reasonable. On the diagnostic split, the standardized residual
\((C-\hat f_3(S))/\hat\sigma(S)\) has empirical standard deviation \(0.994\); the corresponding one- and two-standard-deviation coverages are \(71.3\%\) and \(94.4\%\), close to the Gaussian reference values.

To test the dynamical effect of this learned scale, we use a time-discrete
random-force closure
\begin{equation}
C_k^n
=
\hat f_3(S_k^n)
+
\hat\sigma(S_k^n)\xi_k^n,
\qquad
\xi_k^n\sim \mathcal N(0,1),
\label{eq:l96-random-force-closure}
\end{equation}
inside the slow-only Lorenz--96 model. The random force is refreshed once per closure update at the same discrete time step as the slow-only rollout. While the update of Eq.~\eqref{eq:l96-random-force-closure} coincides in form with one Euler--Maruyama step of a learned effective SDE~\citep{dietrich2023learning,zhu2024dyngma}, it is not a discretization of a continuous-time diffusion: we do not identify a self-consistent SDE but freeze the mean closure and learn only the conditional scale, injecting the learned conditional spread of the instantaneous closure forcing.

\begin{figure}[t]
    \centering
    \includegraphics[width=\textwidth]{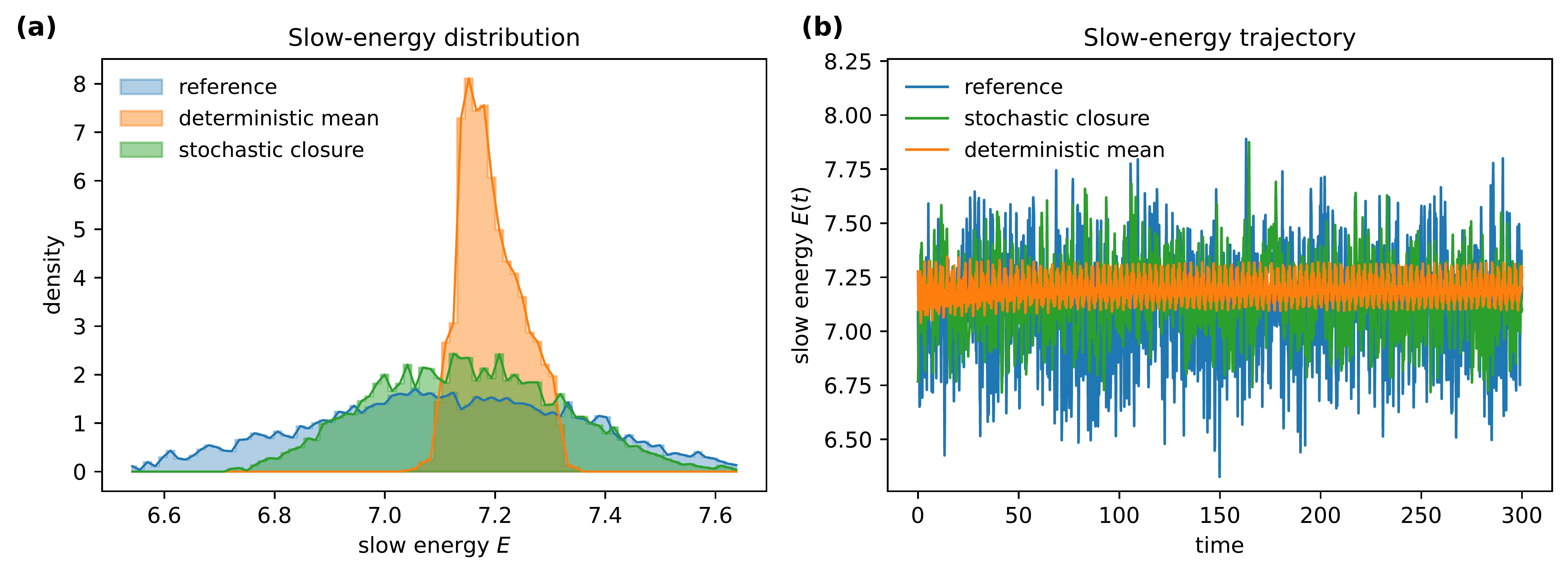}
    \caption{Lorenz--96 slow-energy statistics from independent long
        rollouts of the reference, deterministic-mean, and stochastic closures.
        The density and trajectory panels use one representative stochastic
        rollout; aggregate stochastic ratios are reported in the text over
        \(M=5\) stochastic rollout seeds. (a) Slow-energy densities.
        (b) Slow-energy trajectories.}
    \label{fig:l96-stochastic-summary}
\end{figure}

Figure~\ref{fig:l96-stochastic-summary} compares the full two-scale reference
system, the deterministic slow-only model using \(\hat f_3\), and the
stochastic slow-only model using \eqref{eq:l96-random-force-closure}. The
figure uses one representative stochastic rollout for visualization. Panel (a)
shows the slow-energy density for all three systems: the deterministic mean
closure produces a much narrower energy distribution, while the stochastic
closure broadens it toward the reference. Panel (b) shows representative
slow-energy trajectories, making the same contrast visible in time.
Space--time windows of all three rollouts, confirming that the closed systems remain dynamically active, are displayed in Figure~\ref{fig:l96-spacetime-triptych}.

For quantitative comparison, we run \(M=5\) independent stochastic rollout
seeds while keeping the reference initial condition fixed. Both closed models preserve the one-point variance of \(X_k\) and the mean slow energy reasonably
well:
\[
\frac{\operatorname{Var}_{\mathrm{det}}(X_k)}
{\operatorname{Var}_{\mathrm{ref}}(X_k)}
\approx 1.000,
\qquad
\frac{\operatorname{Var}_{\mathrm{stoch}}(X_k)}
{\operatorname{Var}_{\mathrm{ref}}(X_k)}
=
0.994 \pm 0.001,
\]
\textcolor{black}{where $\operatorname{Var}_{\mathrm{ref}}(\cdot), \operatorname{Var}_{\mathrm{det}}(\cdot), \operatorname{Var}_{\mathrm{stoch}}(\cdot)$ denote the variances of the reference, deterministic, and stochastic rollouts, respectively,} and
\[
\frac{\mathbb E_{\mathrm{det}}[E]}
{\mathbb E_{\mathrm{ref}}[E]}
\approx 1.011,
\qquad
\frac{\mathbb E_{\mathrm{stoch}}[E]}
{\mathbb E_{\mathrm{ref}}[E]}
=
1.005 \pm 0.001 .
\]
The main difference is in the collective fluctuation statistic:
\[
\frac{\operatorname{Var}_{\mathrm{det}}(E)}
{\operatorname{Var}_{\mathrm{ref}}(E)}
\approx 0.055,
\qquad
\frac{\operatorname{Var}_{\mathrm{stoch}}(E)}
{\operatorname{Var}_{\mathrm{ref}}(E)}
=
0.543 \pm 0.034 .
\]
Here the uncertainty is one sample standard deviation over the \(M=5\)
stochastic rollout seeds. Thus the Gaussian scale model raises the recovered
slow-energy variance from about \(5.5\%\) to about \(54\%\) of the reference
value. It does not fully recover the reference statistics, however. This is
expected: \textcolor{black}{it injects a unimodal Gaussian conditional scale, temporally independent and
around a frozen mean, and so cannot represent skewness, multimodality, or
temporal correlation in the conditional forcing. More expressive distributional
models---for example mixtures of Gaussians, diffusion models, or stochastic
closures with temporal structure---may capture these features and improve the
recovery where the task requires it.}

\textcolor{black}{\subsection{Diagnosing a case with weaker scale separation}}
\label{subsec:l96-weaker-separation}

\textcolor{black}{Now we apply the diagnosis at a different parameter setting of the Lorenz--96 system, \(c=b=5\), a regime of weaker scale separation. This creates a more challenging closure problem, as the fast variables carry longer memory and couple more strongly to the slow variables. All other settings, including the stencil \(S_{t,k}=(X_{k-1},X_k,X_{k+1})\), the polynomial family, the diagnostic procedure, and the tolerance \(\varepsilon=0.01\), are unchanged, so that any difference in the diagnosis is attributable to the scale separation alone.}

\textcolor{black}{The diagnostic outcome is qualitatively the same but quantitatively shifted.
As the polynomial degree increases, the largest out-of-basis effect size again
falls below the tolerance, so the closure reaches the conditional-mean barrier;
the crossing now occurs at \(p=4\) rather than at \(p=3\). The explained-variance
ceiling, however, is markedly higher: the normalized aleatoric floor rises from
\(1-R^2\approx 0.26\) at \(c=b=10\) to \(1-R^2\approx 0.46\) at \(c=b=5\)
(Table~\ref{tab:l96-scale-separation}), leaving a substantially larger fraction of the conditional variability irreducible with respect to the chosen stencil.}

\begin{table}[t]
\centering
\caption{Diagnosis under two scale-separation strengths. The selected degree
\(p\) is the smallest degree at which the largest out-of-basis effect size
falls below \(\varepsilon=0.01\). The normalized aleatoric floor \(1-R^2\) is
reported on the diagnostic split at the selected degree. Weaker separation
yields a higher floor.}
\label{tab:l96-scale-separation}
\begin{tabular}{lccc}
\hline
\(c=b\) & Scale separation & Barrier degree \(p\) & Floor \(1-R^2\) \\
\hline
\(10\) & standard         & \(3\) & \(0.26\) \\
\(5\)  & weaker  & \(4\) & \(0.46\) \\
\hline
\end{tabular}
\end{table}

\textcolor{black}{This comparison also shows how the diagnosis informs the next modeling step.
Because the uniformly small effect sizes certify that the \(46\%\) floor at
\(c=b=5\) is not an artifact of an inexpressive mean model, it is genuinely
irreducible \emph{with respect to the three-point stencil}. Reaching the barrier certifies that no deterministic model on the chosen descriptor can do better; it does not certify that the
descriptor is itself a good conditioning variable, and the height of the floor speaks precisely to that question. The choice of descriptor therefore sits above
the regime classification: changing it redefines the conditional law and hence
the entire diagnosis, rather than acting within a fixed regime. A high floor is
the signal that this prior choice is worth revisiting---a more informative
descriptor, such as a wider stencil or a delay embedding, could lower the floor
and recover as deterministic structure what would otherwise be carried by a
stochastic model. It is therefore the sensible first step to test a richer descriptor before committing to a stochastic model of the remaining spread. Constructing
such descriptors and quantifying the resulting gain are beyond the scope of this diagnostic study; our point is that the diagnosis identifies which direction of improvement is warranted.}
\newline

\section{Conclusions}\label{sec:conclusion}

This paper has developed the conditional-mean barrier as a practical diagnostic for scientific machine-learning surrogates trained by squared loss. \textcolor{black}{Taking as its starting point the classical fact that squared loss targets the conditional mean rather than the full conditional law, the paper turns this distinction into an operational diagnostic for finite data.} Residual-feature orthogonality, effect-size diagnostics, and the explained-variance ceiling provide tools for recognizing when this target has been reached, while the variance-suppression decomposition explains why adding latent randomness to a squared-loss-trained predictor does not by itself overcome the barrier.

The resulting decision procedure separates two situations that can otherwise
look similar. If the residual still carries detectable and practically important
structure in the chosen input variables, the deterministic surrogate should be
refined. If no such structure remains but the residual variability is still too
large for the downstream task, then the model has reached the conditional-mean
barrier relative to the chosen information \(X\). At that point, the modeling
target must move from a point predictor to an appropriate feature of
\(P(Y\mid X)\).

The numerical studies illustrate this distinction in both a controlled
two-branch law and a two-scale Lorenz--96 closure problem. In the latter case,
a deterministic closure selected by the diagnostic preserves simple one-point
statistics but suppresses collective slow-energy fluctuations in autonomous
rollout; a minimal likelihood-based stochastic-scale intervention recovers a
substantial part of this missing variability. These results emphasize that the
loss function is not a technical afterthought: it determines what object the
surrogate learns, and therefore which scientific quantities it can represent.

\textcolor{black}{Although demonstrated here with polynomial regressors, the diagnostic is not
tied to that choice. It uses only the fitted residual \(Y-f_\theta(X)\) and a
family of probes, so the effect size, significance test, and explained-variance
ceiling can be computed identically for a neural surrogate or a neural operator. The
one component that must be adapted is the capacity path: where the polynomial
degree gives a natural ladder for refining the deterministic model, a neural
model is refined instead by widening or deepening the network, training longer,
or changing architecture, with the same residual diagnostic applied at each
level. For a neural operator, whose inputs and outputs are functions, the probes
can be functionals of the input---spectral coefficients or learned latent
codes---and the residual is assessed in the output function space. Implementing
and stress-testing the diagnostic for such black-box models is an interesting direction for future work.}

\appendix
\section{Space--time rollouts for the Lorenz--96 closures}
\label{sec:supp-spacetime}

Here, in Fig. A1 we display the space--time windows of the resolved variables $X_k$ for the three Lorenz--96 systems discussed in the main article: the full two-scale reference system, the deterministic-mean closed slow-only system, and the stochastic closed slow-only system. These windows accompany the downstream slow-energy statistics reported in the main text; their role is to confirm that each closed system remains dynamically active rather than collapsing to a fixed state.

\begin{figure}[htbp]
    \centering
    \includegraphics[width=0.8\textwidth]{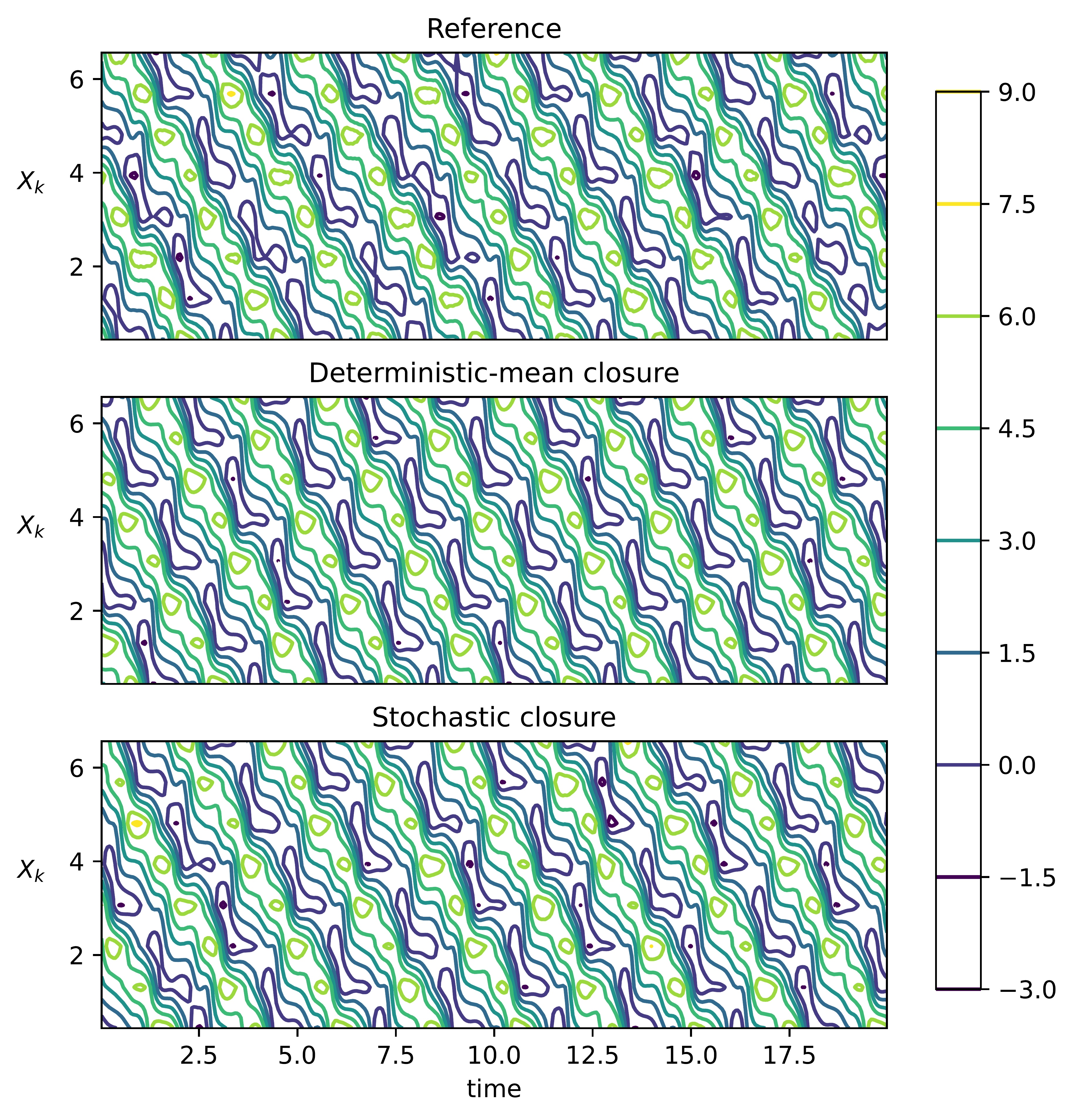}
    \caption{Short space--time windows of the resolved variables $X_k$ from
    independent long rollouts, on a shared color scale: the full two-scale
    reference system (top), the deterministic-mean closed slow-only system
    (middle), and the stochastic closed slow-only system (bottom). All three
    remain dynamically active rather than collapsing to a fixed state.}
    \label{fig:l96-spacetime-triptych}
\end{figure}

\section*{Data and code availability}
Data and code are available at \url{https://github.com/junfeng-chen/conditional_mean_barrier}.

\section*{Declaration of competing interest}
The author declares that they have no known competing financial interests or personal relationships that could have appeared to influence the work reported in this paper.

\section*{Acknowledgments}
The author has used AI tools to assist with editing and formatting this manuscript. The author has reviewed and edited the content as needed and takes full responsibility for the final version of the manuscript.

\bibliographystyle{elsarticle-harv}
\bibliography{ref}

\end{document}